\documentclass[10pt,twocolumn,letterpaper]{article}
\usepackage[pagenumbers]{cvpr}

\usepackage[T1]{fontenc}
\usepackage{microtype}
\usepackage{placeins}

\definecolor{vmcblue}{HTML}{0072B2}
\definecolor{vmcorange}{HTML}{E69F00}
\definecolor{vmcgreen}{HTML}{009E73}
\definecolor{vmcred}{HTML}{C73E4D}
\usepackage{array}

\AtBeginDocument{\raggedbottom}

\usepackage{dblfloatfix}

\definecolor{cvprblue}{rgb}{0.21,0.49,0.74}
\usepackage{listings}
\usepackage{hyperref}
\hypersetup{colorlinks=true,allcolors=cvprblue}

\title{MMMem: Self-Improving Agents with Task-Aware Multimodal Experience Consolidation}
\author{%
  Yuxuan Chen\textsuperscript{1,*}\quad
  Changwei Lv\textsuperscript{2,*}\quad
  Yunfei Li\textsuperscript{3}\quad
  Zhongjing Du\textsuperscript{1}\quad
  Kehui Tan\textsuperscript{1}\quad
  Zijun Liao\textsuperscript{1}\\
  Ruochong Zheng\textsuperscript{1}\quad
  Hao Chen\textsuperscript{1}\quad
  Can Deng\textsuperscript{4}\quad
  Zheni Zeng\textsuperscript{3}\quad
  Chang Liu\textsuperscript{4}\quad
  Jie Chen\textsuperscript{1,\S}\\[4pt]
  {\small\textsuperscript{1}Peking University\quad
  \textsuperscript{2}Clemson University\quad
  \textsuperscript{3}Nanjing University\quad
  \textsuperscript{4}Tsinghua University}\\[2pt]
  {\small\textsuperscript{*}Equal contribution\quad
  \textsuperscript{\S}Corresponding author}%
}

\hypersetup{
  pdftitle={MMMem: Self-Improving Agents with Task-Aware Multimodal Experience Consolidation},
  pdfauthor={Yuxuan Chen, Changwei Lv, Yunfei Li, Zhongjing Du, Kehui Tan, Zijun Liao, Ruochong Zheng, Hao Chen, Can Deng, Zheni Zeng, Chang Liu, Jie Chen}
}

\begin{document}
\maketitle

\begin{abstract}

AI agents accumulate conversations, images, and documents through long-term interactions with users. Answering later questions often requires connecting evidence across these records, yet the resulting understanding is rarely retained, allowing mistakes to recur on related questions. Existing multimodal memory methods largely focus on storing and retrieving records, with limited mechanisms for learning from completed tasks and consolidating task-specific experience. We therefore propose \textbf{MMMem}, a task-aware multimodal memory framework that extracts and consolidates experience from completed tasks through four complementary strategies. First, source-linked multi-view memory organizes original text, images, and rendered pages alongside generated scene images that combine clues across records for retrieval. Second, source-generated tasks support knowledge alignment at two levels: warm-up consolidates source knowledge into task-relevant experiences in external memory, while self-training adapts the model's parametric knowledge to task demands. Third, the same vision-language model distills completed tasks into experiences that record the task context, conclusion, and supporting source links. Finally, task-aware retrieval recalls source views and relevant experiences for answering and verification, while source-supported reuse guides long-term retention. Across three memory benchmarks and two vision-language readers, MMMem outperforms eight baselines. With Qwen3.5-9B, MMMem achieves relative gains of 36.9\%, 17.6\%, and 15.2\% over the best-performing baselines on LoCoMo, Mem-Gallery, and LooGLE v2, respectively. Ablations show consistent gains from training and memory warm-up, while case studies illustrate how individual components work in practice.

\end{abstract}

\section{Introduction}
\label{sec:introduction}

\begin{figure*}[t]
  \centering
  \includegraphics[width=\textwidth]{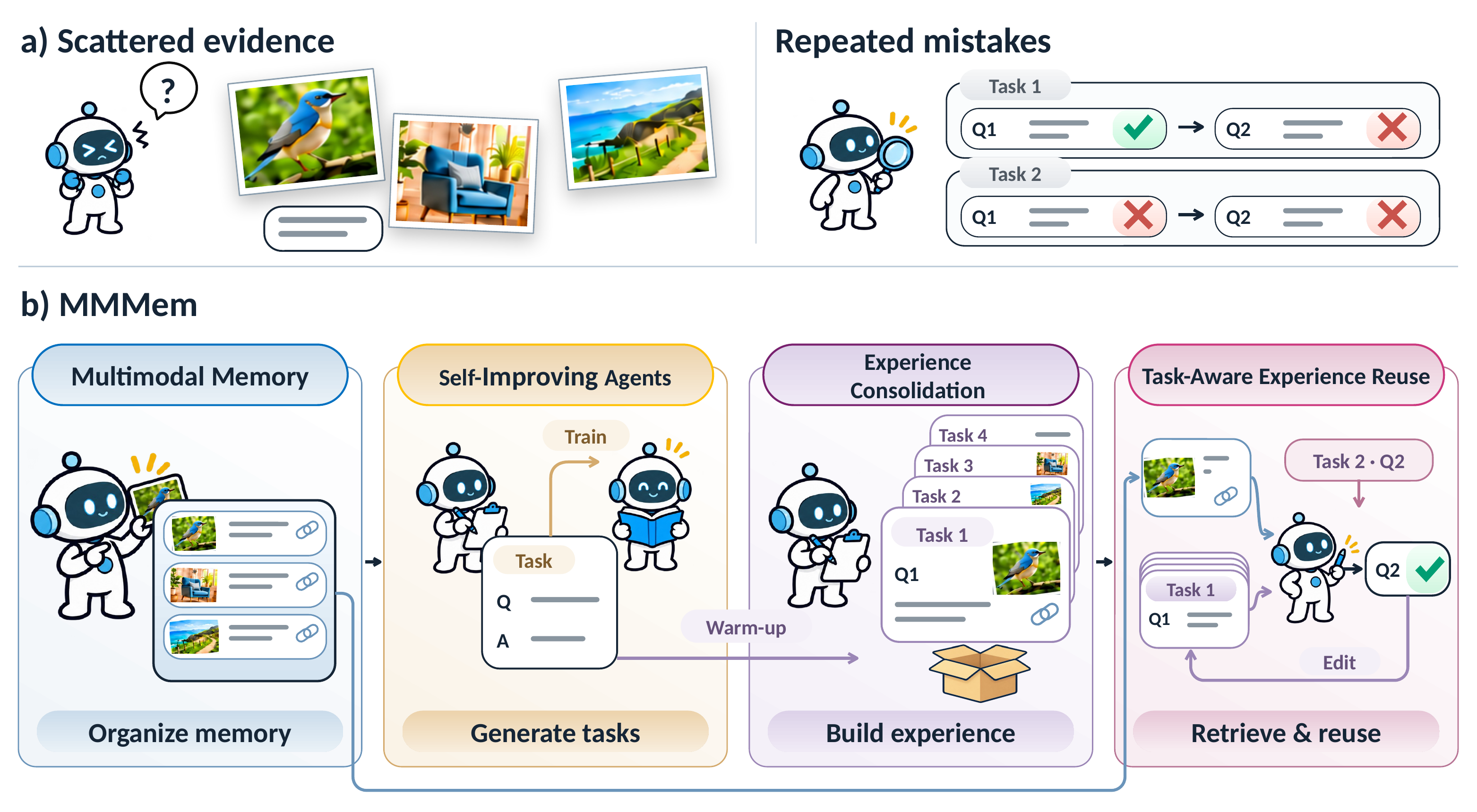}
  \caption{\textbf{Motivation and overview of MMMem.}
  (a) Without task-level memory, the evidence for a question stays scattered across records, and the agent can repeat mistakes on related questions.
  (b) MMMem organizes multimodal memory, generates tasks for warm-up and training, turns solved tasks into experiences, and retrieves and reuses them on later tasks.}
  \label{fig:teaser}
\end{figure*}

AI assistants that support users over time accumulate conversations, photographs, and documents across interactions. Later requests may depend on information spread across these records, requiring the assistant to connect complementary cues from different modalities. For example, identifying an item in a photograph may require connecting its visual appearance with a name mentioned in an earlier conversation. The agent must determine which records belong together, what they jointly establish, and where each conclusion comes from. These connections emerge during task solving, but a memory that stores only the original records leaves this understanding to be reconstructed for later questions, and the agent may repeat the same mistakes on related questions (Fig.~\ref{fig:teaser}a). Preserving this understanding together with its supporting sources would allow completed tasks to improve how memory is organized and used.

The current task determines which evidence to prioritize and which earlier conclusions are relevant. An identity question may rely on names in a dialogue, whereas a question about appearance may require visual details. An entity label stored without its task context and visual grounding can be wrongly transferred to another image when questions are similar. We therefore use tasks as the organizing unit of multimodal memory. Each resulting \emph{experience} records the task context, conclusion, and supporting source links, making explicit what was learned and which evidence supports it.

Existing memory methods retrieve relevant context \citep{lewis2020rag,asai2024selfrag,xu2024recomp} and organize evolving records \citep{chhikara2025mem0,xu2025amem}, while visual memory methods \citep{li2026ocrmemory,shi2026memocr} and experience-based agents \citep{shinn2023reflexion,zhao2024expel,wang2024awm,ouyang2026reasoningbank} extend how histories are represented and reused. These approaches offer complementary capabilities; the challenge is to coordinate knowledge consolidation in external memory with parametric adaptation, allowing agents to learn from multimodal experience across tasks. MMMem combines external memory with model adaptation in a task-aware framework, enabling agents to learn from multimodal experience and to consolidate, reuse, and retain what they learn as they solve tasks. Section~\ref{sec:method} presents the method in detail.

MMMem improves both the organization of multimodal knowledge and the agent's ability to use it (Figs.~\ref{fig:teaser}b and~\ref{fig:pipeline}). Its source-linked memory retains original records alongside generated scene views that combine clues across records, providing complementary representations for task-aware retrieval. Source-generated tasks connect these two forms of improvement: warm-up organizes source knowledge into retrievable experience, while self-training adapts the model to interpret evidence and consolidate experience. Completed tasks yield experiences that preserve their conclusions and the sources supporting them. Later reuse, when supported by the original evidence, guides long-term retention of these experiences.

Experiments on conversational memory, long-document question answering, and multimodal understanding \citep{maharana2024locomo,he2025looglev2,bei2026memgallery} show that MMMem consistently outperforms the compared memory baselines across reader models and memory tasks. These results demonstrate the framework's effectiveness in using accumulated multimodal information for retrieval and reasoning, supporting its applicability across models and tasks. Ablation studies further examine the contributions of self-training, task warm-up, experience memory, and generated views, and compare MMMem with applying the same trained reader to dense retrieval alone.

Our main contributions are as follows: (1) We propose MMMem, a framework for task-aware multimodal experience consolidation that turns completed tasks into source-linked experiences and guides their retention through evidence-supported reuse; (2) We develop complementary warm-up and self-training strategies that use source-generated tasks to organize source knowledge into reusable experience and adapt the model for evidence-based answering and experience consolidation; (3) We evaluate MMMem across three memory benchmarks and two vision-language models, using ablations and case studies to examine how its core mechanisms support task performance.

\section{Related Work}
\label{sec:related_work}

\paragraph{Multimodal Memory.}
Multimodal memory aims to preserve information across text, images, and document layouts for later use. Building on retrieval-augmented generation \citep{lewis2020rag}, existing methods retrieve document images directly \citep{faysse2025colpali,yu2025visrag} or render interaction histories into images to recover or emphasize relevant evidence \citep{li2026ocrmemory,shi2026memocr}. Beyond representation, recent work studies how retrieved evidence is delivered to the model, preserving its modality, identity, and timing \citep{jiang2026delivermem} or selecting among textual, visual, and dual-channel representations \citep{gao2026memo}. MMMem links original records and generated views through source identifiers, allowing task-relevant evidence to be combined across representations and traced back to its sources.

\paragraph{Memory Editing.}
As interactions accumulate, memory systems must update what they retain. Existing methods extract and maintain facts from conversations \citep{chhikara2025mem0} or organize memory notes through links that evolve with new information \citep{xu2025amem}. Task execution provides another source of memory: verbal reflections and insights can be retained for later decisions \citep{shinn2023reflexion,zhao2024expel}, while execution trajectories can be distilled into reusable workflows and reasoning strategies \citep{wang2024awm,ouyang2026reasoningbank}. MMMem records multimodal question-answering experiences as tasks, conclusions, and supporting source links. These experiences are immediately retrievable, while later reuse supported by source evidence guides long-term retention. The supplementary material compares their memory content, retrieval and reading mechanisms, updates, and trained components.

\paragraph{Parametric Memory.}
Parametric memory complements external storage by encoding knowledge in model weights \citep{petroni2019lama}. Recent approaches consolidate short-term knowledge through distillation \citep{behrouz2026sleep} or incorporate interaction experience through online updates and experience distillation \citep{ren2026tmem,ye2026oel}. Others coordinate episodic storage with the gradual consolidation of recurring patterns into parameters \citep{xia2026unimem}. Parameter learning also improves how models use external evidence, from deciding when to retrieve and assessing answers \citep{asai2024selfrag} to identifying relevant information amid distractors \citep{zhang2024raft}. MMMem trains a shared model for evidence-based answering and experience consolidation, connecting parameter learning with the use and construction of external memory.

\paragraph{Self-Improvement.}
The effectiveness of parameter adaptation depends on the training signals available to the model. Self-improvement methods construct these signals from generated outputs and feedback. Some methods improve solutions to existing tasks by training on rationales that yield correct answers \citep{zelikman2022star} or combining generated samples with offline reinforcement learning \citep{gulcehre2023rest}. In vision-language learning, verified programmatic reasoning can provide supervision for model training \citep{hu2024vpd}. Another line of work generates question-answer pairs from knowledge sources through multi-grained self-annotation and iteratively adapts models through self-verification \citep{zeng2025kbalign}. MMMem uses tasks generated from source materials for memory warm-up and self-training, accumulating external experiences with fixed model parameters and training the shared model for answering and experience writing on separate training sources.

\section{Method}
\label{sec:method}

\begin{figure*}[t]
  \centering
  \includegraphics[width=\textwidth]{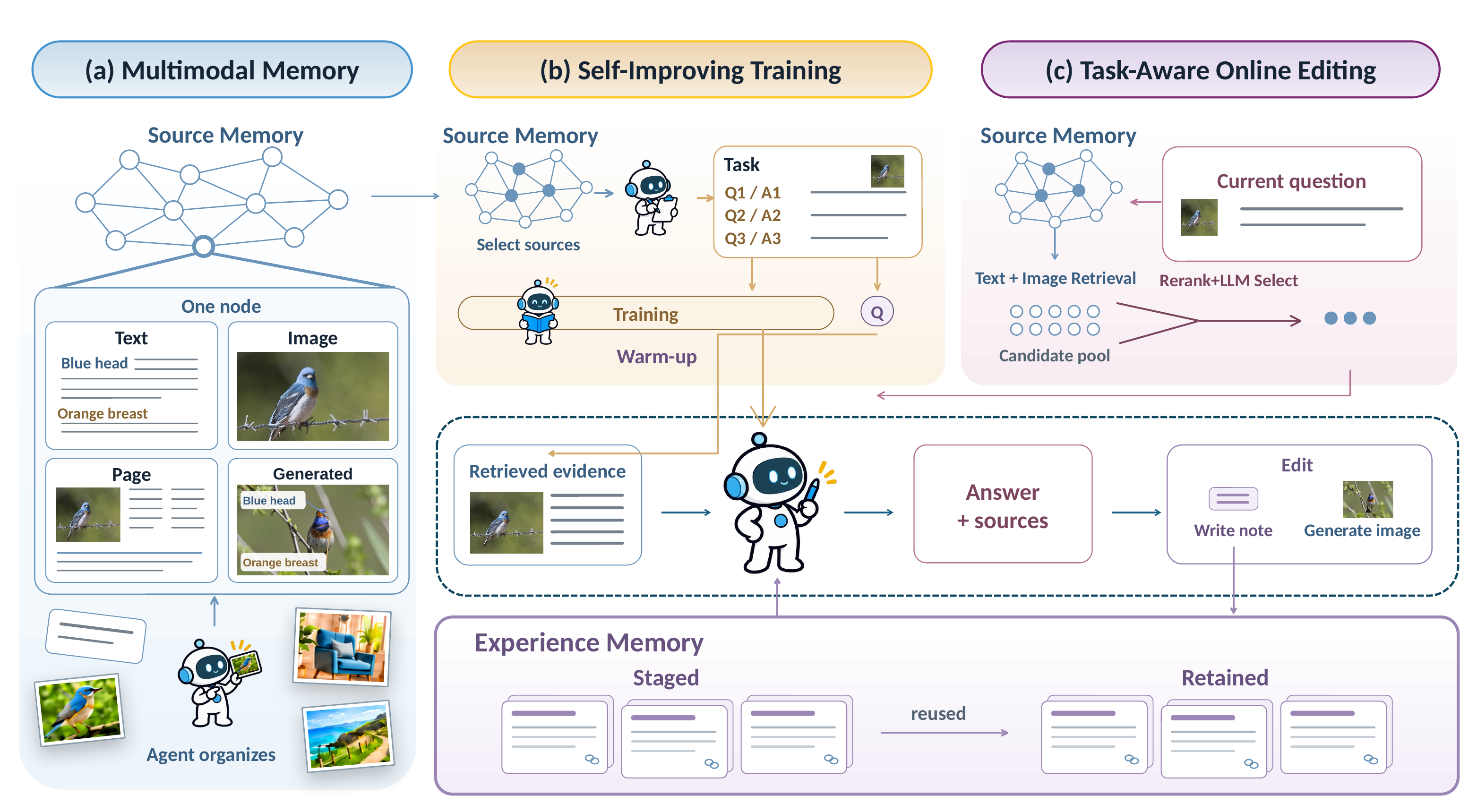}
  \caption{\textbf{MMMem framework.}
  (a) Each memory node holds text, image, page, and generated views of its source records.
  (b) Tasks generated from the source memory are used to train the reader offline and to warm up the memory.
  (c) For the current question, text and image retrieval followed by reranking selects the evidence. The agent answers with its sources, writes a note, and may generate an image; the resulting experience moves from staged to retained after repeated, source-supported reuse.}
  \label{fig:pipeline}
\end{figure*}

MMMem combines complementary views of multimodal records with experiences distilled from completed tasks. Tasks generated from source material support memory warm-up with fixed model parameters and self-training on separate training sources. For each question, a vision-language model answers using retrieved evidence and relevant experience, then writes a new experience in a second call. We describe these three components in turn (Fig.~\ref{fig:pipeline}): multimodal memory (Section~\ref{sec:views}), self-improving training (Section~\ref{sec:selfgen}), and task-aware online editing (Section~\ref{sec:online}).

\subsection{Multimodal Memory}
\label{sec:views}

MMMem organizes records with persistent source IDs into local and cross-record memory nodes (Fig.~\ref{fig:pipeline}a). Consecutive turns within a session form segment nodes, and a session node links these segments. Entity nodes collect records sharing a recurring capitalized name across sessions; a single timeline node gathers records containing temporal expressions within a memory unit. A record therefore belongs to one segment and can also contribute to entity or timeline nodes. These groupings provide local context and access to information distributed across interactions.

Node contents are exposed through original text and images, page images with OCR text, and optional generated scenes. Scene generation initially targets entity and timeline nodes; later usage can trigger generation for additional nodes. The model reads all records associated with the node and selects the facts to depict in one scene.

The selected facts retain explicit source IDs and are summarized into a scene description for image generation. Fact IDs are checked against the node's records, and a read-back check compares the generated image's people and setting with its specification. Generated views retain the provenance of the cited facts.

All image views are indexed for visual retrieval, so a match to an original, page, or generated image can retrieve its node. The answering model then receives the selected node's text and original images. Citations follow the retained source IDs back to the records, preserving the connection between the representation used for retrieval and the evidence used for answering.

\subsection{Self-Improving Training}
\label{sec:selfgen}

Source-generated tasks address the gap between available knowledge and task requirements at two complementary levels (Fig.~\ref{fig:pipeline}b). Warm-up aligns external memory with task requirements by consolidating source knowledge into retrievable experiences. Self-training provides internal alignment by adapting the model's parametric knowledge for evidence-based answering and experience consolidation. Each task is constructed from selected memory nodes and includes a question, a target answer, and supporting record IDs.

The generation prompts cover single-record facts, temporal and multi-record reasoning, and questions whose answers are absent from the records. Alongside independent questions, related sequences revisit the same fact through paraphrases, request different information from the same sources, or introduce a nearby unsupported query. This gives generated tasks a role in both initial learning and subsequent experience reuse.

\paragraph{Warm-Up.}
Before user queries arrive, MMMem runs generated questions through the answering and experience-writing procedure in Section~\ref{sec:online}. The model receives the generated question, any answer options, and retrieved source evidence, then produces its own answer. The model parameters remain fixed, while the resulting experiences enter the staging area and are immediately retrievable. This creates an initial set of task-relevant conclusions and evidence associations for later questions.

\paragraph{Self-Training.}
Offline self-training uses source-generated tasks from five benchmark domains, with training units disjoint from evaluation units. The MiniCPM recipe includes 54 image-based QA examples generated by Qwen3.5-9B. Tasks generated from these sources yield two supervision targets: answers with supporting evidence, and experience summaries containing the task context, conclusion, and source links. The two objectives train a shared vision-language model by updating its LoRA parameters $\theta$ \citep{hu2022lora}, with the backbone weights fixed. Answer examples supervise evidence use, whereas summary examples supervise the organization of task-specific knowledge into reusable experience.

\begin{align}
  \mathcal{L}(\theta)
    &= \mathcal{L}_{\mathrm{answer}}(\theta)
       + \mathcal{L}_{\mathrm{summary}}(\theta),
       \label{eq:loss}\\
  \mathcal{L}_k(\theta)
    &= -\!\sum_{(x,y)\in\mathcal{D}_k}
       \sum_{t=1}^{|y|}
       \log p_\theta(y_t\mid x,y_{<t}),\notag
\end{align}

Here, $k\in\{\mathrm{answer},\mathrm{summary}\}$ indexes the two example sets $\mathcal{D}_k$, $x$ is the input context, and $y$ is the target sequence. Loss is computed over target tokens only. During inference, answering and experience writing share the trained parameters, which remain fixed as external memory is updated.

\subsection{Task-Aware Online Editing}
\label{sec:editing}
\label{sec:answering}
\label{sec:online}

\paragraph{Retrieval and Answering.}
For a task $q$, MMMem retrieves candidate views through text and visual search and reranks them to assemble evidence $\mathcal{V}_q$ within the source-view budget (Fig.~\ref{fig:pipeline}c). Retrieved records supply current evidence, while recalled experiences provide prior task interpretations and conclusions for the model to check against that evidence. The shared vision-language model combines this evidence with a recalled experience $\tilde e_q$ and any query image to produce the answer,

\begin{equation}
  \hat c = R_\theta(q,\mathcal{V}_q,\tilde e_q).
  \label{eq:read}
\end{equation}

\paragraph{Experience Consolidation.}
\label{sec:retention}
\label{sec:consolidation}
To preserve what was established during answering, a second call to the same model writes a structured experience,

\begin{equation}
  e=(\tau,c,\mathcal{L},\sigma)
    = W_\theta(q,\hat c,\mathcal{V}_q),
  \label{eq:experience}
\end{equation}

where $\tau$ describes the task, $c$ is the distilled conclusion, $\mathcal{L}$ contains supporting source links, and $\sigma$ is the retention state, initially $\mathrm{staged}$. The record preserves the relationship between a task's conclusion and the particular evidence selected to support it. Both $R_\theta$ and $W_\theta$ use the same fixed parameters under different prompts. On its first adoption by a later task, an experience may receive a generated image sharing its task description, conclusion, and source links. This derived view accompanies the experience when recalled.

\paragraph{Experience Recall.}
New experiences are immediately retrievable. For each task, staged experiences are ranked by question-text relevance, and the top entry occupies a dedicated slot $\tilde e_q$ outside the source-view budget. This recall route remains active in all settings. Staging makes new experience available immediately while leaving its longer-term retention to be determined by subsequent use.

\paragraph{Reuse-Guided Retention.}
An experience is credited as adopted when it reaches the reader and the answer cites at least one linked record that supports the answer. This links retention to continued usefulness on later questions and support from the original records. Rephrasings of the same question contribute at most two adoptions. Every 8 tasks, or when the 32-entry staging area is full, experiences with at least two adoptions are retained and indexed with ordinary memory nodes. At each check, staged experiences with no adoption for 32 tasks expire. Promotion changes only $\sigma$, preserving the task description, conclusion, and source links.

\begin{table*}[!b]
  \centering
  \small
  \setlength{\tabcolsep}{2.8pt}
  \renewcommand{\arraystretch}{1.08}
  \begin{tabular}{@{}l*{5}{c}@{\hspace{10pt}}*{5}{c}@{}}
    \toprule
    & \multicolumn{5}{c}{\textbf{Qwen3.5-9B}} & \multicolumn{5}{c}{\textbf{MiniCPM-V 4.5}} \\
    \cmidrule(lr){2-6}\cmidrule(lr){7-11}
    & LoCoMo~\cite{maharana2024locomo} & \multicolumn{3}{c}{Mem-Gallery~\cite{bei2026memgallery}} & LooGLE v2~\cite{he2025looglev2}
    & LoCoMo & \multicolumn{3}{c}{Mem-Gallery} & LooGLE v2 \\
    \cmidrule(lr){2-2}\cmidrule(lr){3-5}\cmidrule(lr){6-6}
    \cmidrule(lr){7-7}\cmidrule(lr){8-10}\cmidrule(lr){11-11}
    Method & F1 & EM & F1 & BLEU-1 & Score & F1 & EM & F1 & BLEU-1 & Score \\
    \midrule
    Direct Answer & 20.97 & 12.28 & 14.30 & 13.12 & 17.87 & 11.80 & 16.49 & 22.21 & 20.04 & 16.58 \\
    Full Context (32K cap) & 35.48 & 33.96 & 51.51 & 46.31 & 24.85 & 7.66 & 3.49 & 8.35 & 6.74 & 10.48 \\
    Recent Context (3K/4K) & 25.72 & 12.81 & 18.42 & 16.42 & 18.55 & 24.66 & 24.91 & 32.15 & 29.64 & 16.67 \\
    Dense RAG~\cite{lewis2020rag} & 45.03 & 37.81 & 55.76 & 51.05 & 26.01 & 43.43 & 46.24 & 64.40 & 59.70 & 21.97 \\
    OCR-Memory~\cite{li2026ocrmemory} & 33.23 & 17.47 & 23.19 & 20.98 & 19.43 & 30.68 & 25.90 & 37.45 & 33.33 & 17.75 \\
    MemOCR~\cite{shi2026memocr} & 47.48 & 36.92 & 48.97 & 45.23 & 21.16 & 48.09 & 35.39 & 51.06 & 45.60 & 20.49 \\
    Mem0~\cite{chhikara2025mem0} & 31.84 & 21.59 & 31.99 & 27.49 & 19.56 & 30.74 & 30.20 & 42.87 & 37.74 & 18.33 \\
    A-Mem~\cite{xu2025amem} & 48.09 & 47.58 & 65.63 & 60.71 & 22.05 & 23.98 & 31.45 & 47.31 & 43.03 & 12.05 \\
    \midrule
    \textbf{Ours}& \textbf{65.84} & \textbf{61.56} & \textbf{77.21} & \textbf{71.65} & \textbf{29.97} & \textbf{56.40} & \textbf{53.23} & \textbf{70.27} & \textbf{65.23} & \textbf{24.67} \\
    \bottomrule
  \end{tabular}
  \caption{\textbf{Main evaluation results across three memory benchmarks.} We compare MMMem against eight representative baselines under shared Qwen3.5-9B and MiniCPM-V 4.5 reader backbones across long-term conversations (LoCoMo), multimodal interactions (Mem-Gallery), and long documents (LooGLE v2).}
  \label{tab:main}
\end{table*}

\section{Experiments}
\label{sec:experiments}

We evaluate MMMem across three memory benchmarks and two vision-language models. The experiments compare performance with existing methods and examine the contributions of self-training, memory organization, and individual components (Sections~\ref{sec:results} and~\ref{sec:ablation}). Case studies illustrate how the mechanisms affect evidence retrieval and experience reuse (Section~\ref{sec:case_study}).

\subsection{Experimental Setup}
\label{sec:setup}

\paragraph{Benchmarks.}
We evaluate on held-out subsets of three benchmarks. LoCoMo~\citep{maharana2024locomo} tests recall and reasoning over multi-session conversations; Mem-Gallery~\citep{bei2026memgallery} tests image--dialogue associations; and LooGLE v2~\citep{he2025looglev2} tests evidence integration across long documents. Each question accesses only its associated dialogue, scenario, or document. Evaluation-set statistics are provided in the supplementary material.

\paragraph{Evaluation Metrics.}
We report token F1 on LoCoMo; EM, token F1, and BLEU-1 on Mem-Gallery; and the official aggregate score on LooGLE v2, including partial credit for version-control questions. Unanswerable Mem-Gallery questions expect ``Not mentioned.'' Failed or malformed outputs remain in the evaluation denominator.

\paragraph{Readers.}
We use Qwen3.5-9B and MiniCPM-V 4.5 (8.7B parameters) as reader backbones. Comparisons within each reader setting share the same base backbone. MMMem is trained separately for each reader.

\paragraph{Baselines.}
We compare eight baselines in four groups. Context baselines include Direct Answer, which uses only the question and any query image; Full Context, which supplies up to 32K history tokens; and Recent Context, which retains 3K tokens on LoCoMo and 4K on the other benchmarks. Dense RAG~\citep{lewis2020rag} retrieves history passages through semantic matching. Visual memory baselines, OCR-Memory~\citep{li2026ocrmemory} and MemOCR~\citep{shi2026memocr}, provide history as rendered pages. Agent memory baselines, Mem0~\citep{chhikara2025mem0} and A-Mem~\citep{xu2025amem}, organize history through factual updates and evolving linked notes, respectively.

Prior memory methods incorporate training at different stages, including retrieval, reading/writing, and memory management, as summarized in the supplementary material. Our ablations compare MMMem with its counterpart without self-training and with Dense RAG using the same trained reader, separating the contributions of model adaptation and the memory workflow.

\paragraph{Memory and Retrieval.}
MMMem uses source-view budgets of 3,000 tokens on LoCoMo and 4,000 on Mem-Gallery and LooGLE v2, with zero-shot reranking by Qwen3-Reranker-0.6B. LoCoMo and Mem-Gallery use text and visual retrieval; retrieval on LooGLE v2 is predominantly text-based. Gallery images are indexed through captions and the visual route, while query images support retrieval and are supplied to the reader. The top-ranked experience occupies a separate slot outside the source budget.

Baselines do not write evaluation questions or model answers into memory for subsequent questions. MMMem processes questions in dataset order within each dialogue, scenario, or document, accumulating experiences from its own answers and cited source records. Experience memory resets between units, and each unit begins with unscored warm-up tasks generated from its own sources. Model parameters remain fixed throughout warm-up and evaluation. Experience writing uses no reference answers or scoring feedback. Benchmark reference answers, annotated clues, and QA tags are excluded from retrieval.

\paragraph{Self-Training.}
Self-training uses tasks generated from separate training units of LoCoMo, Mem-Gallery, LooGLE v2, LVBench, and OVO-Bench. These units are disjoint from evaluation; LVBench and OVO-Bench contribute training data only. Official training-unit question texts guide question style and abstention formatting, without their reference answers. Evaluation-unit questions and answers are excluded from these procedures. The MiniCPM training set includes 54 image-based QA examples generated by Qwen3.5-9B. Each reader is trained for one epoch using attention-projection LoRA adapters and AdamW. Hyperparameters and data-generation details are provided in the supplementary material.

\begin{table*}[t]
\centering\small
\setlength{\tabcolsep}{3pt}
\begin{tabular}{lcccccc}
\toprule
& \multicolumn{3}{c}{\textbf{Qwen3.5-9B}} & \multicolumn{3}{c}{\textbf{MiniCPM-V 4.5}} \\
\cmidrule(lr){2-4}\cmidrule(lr){5-7}
& LoCoMo~\cite{maharana2024locomo} & Mem-Gallery~\cite{bei2026memgallery} & LooGLE v2~\cite{he2025looglev2} & LoCoMo & Mem-Gallery & LooGLE v2 \\
Setting & F1 & F1 & Score & F1 & F1 & Score \\
\midrule
Ours & \textbf{65.84} & \textbf{77.21} & \textbf{29.97} & \textbf{56.40} & \textbf{70.27} & \textbf{24.67} \\
\midrule
Only Training & 60.19 ($-5.65$) & 70.87 ($-6.34$) & 26.11 ($-3.86$) & 50.70 ($-5.70$) & 65.82 ($-4.45$) & 23.45 ($-1.22$) \\
w/o Training & 54.23 ($-11.61$) & 66.21 ($-11.00$) & 24.51 ($-5.46$) & 45.96 ($-10.44$) & 66.11 ($-4.16$) & 23.72 ($-0.95$) \\
w/o Generated views & 64.48 ($-1.36$) & 74.09 ($-3.12$) & --- & 53.68 ($-2.72$) & 69.94 ($-0.33$) & --- \\
w/o Task warm-up & 65.08 ($-0.76$) & 74.81 ($-2.40$) & 23.35 ($-6.62$) & 55.83 ($-0.57$) & 69.09 ($-1.18$) & 22.54 ($-2.13$) \\
w/o Experience & 64.86 ($-0.98$) & 75.12 ($-2.09$) & 24.21 ($-5.76$) & 54.94 ($-1.46$) & 70.09 ($-0.18$) & 22.20 ($-2.47$) \\
\bottomrule
\end{tabular}
\caption{\textbf{Ablation and trained-reader comparison across three benchmarks.} We compare MMMem with component ablations on LoCoMo, Mem-Gallery, and LooGLE v2 with both readers. Dashes indicate unreported results.}
\label{tab:ablation}
\end{table*}
\begin{figure*}[t]
  \centering
  \includegraphics[width=\textwidth]{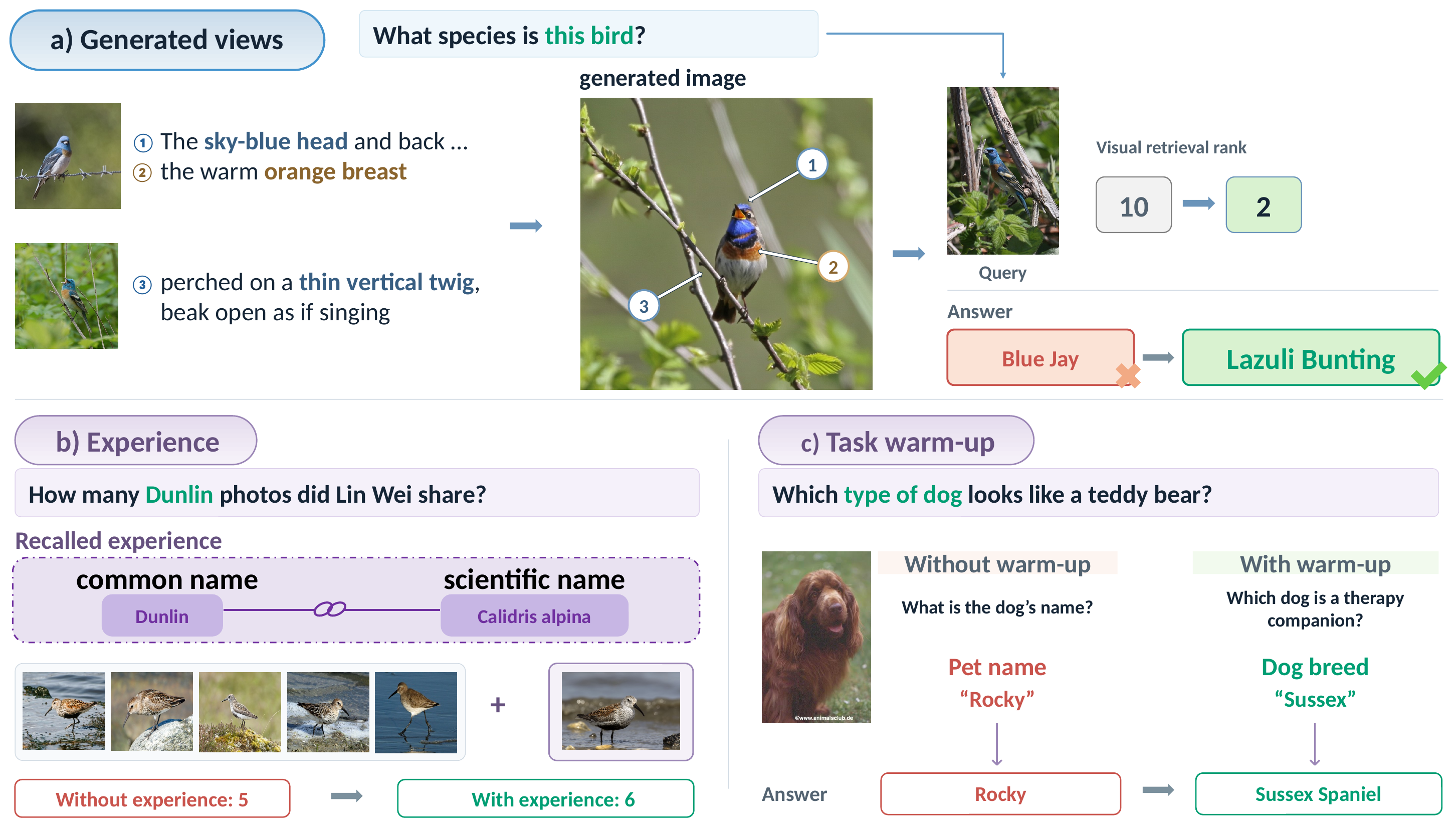}
  \caption{\textbf{Case studies of MMMem.} (a) A generated view combines clues from multiple records for visual retrieval and bird identification. (b) A recalled experience links the common and scientific names of a bird in a photo-counting task. (c) The task warm-up example compares answers to a dog-breed question with and without warm-up.}
  \label{fig:cases}
\end{figure*}

\subsection{Main Results}
\label{sec:results}
\label{sec:main_results}

\paragraph{Overall comparison.}
MMMem achieves the best results on all three benchmarks with both readers (Table~\ref{tab:main}). The gains are particularly large on conversational and multimodal memory: with Qwen3.5-9B, MMMem improves over A-Mem by 17.75 F1 on LoCoMo and 11.58 F1 on Mem-Gallery. MMMem also improves performance with MiniCPM-V~4.5, further supporting its effectiveness.

\paragraph{Comparison with existing methods.}
Dense RAG outperforms Full Context in all six settings. Full Context improves over Direct Answer with Qwen3.5-9B but degrades performance with MiniCPM-V~4.5, highlighting the importance of selecting evidence the reader can use. The effectiveness of memory representations also varies: MemOCR is competitive on LoCoMo but trails Dense RAG on the other benchmarks, while A-Mem's ranking changes across readers. These patterns motivate coordinating memory organization and evidence selection with the reader's ability to interpret the retrieved material.

\paragraph{Differences across benchmarks.}
Gains are smaller on LooGLE v2, where questions require combining evidence across long documents within the source-view token budget. Each document contains about 1.5 scored questions on average, giving the reader fewer opportunities to reuse experience from earlier answers. Warm-up therefore provides an important source of experience before evaluation, consistent with its larger effect on LooGLE v2 in the ablation study.

\paragraph{Effect of self-training.}
The controls in Table~\ref{tab:ablation} show complementary contributions from self-training and the memory workflow. Self-training improves MMMem in all six settings and also improves Dense RAG when the same adapted reader is used. With that reader held fixed, full MMMem further outperforms Dense RAG in every setting. These comparisons show that parameter adaptation accounts for part of the improvement, while the memory workflow supplies additional gains through the organization and use of task-relevant evidence and experience.

\subsection{Ablation Study}
\label{sec:ablation}
\label{sec:analysis}

Table~\ref{tab:ablation} removes components from the full MMMem. Removing training causes the largest drop in four of the six settings; this variant and Only Training are discussed in Sec.~\ref{sec:results}, and we discuss the other three components below.

\emph{Task warm-up} (``MMMem'' \vs ``w/o Task warm-up''). Removing warm-up lowers all six settings, most on LooGLE v2 (6.62 points with Qwen3.5-9B and 2.13 with MiniCPM-V~4.5).
Warm-up writes experiences before the first scored question, which matters most when a unit has few questions: with about 1.5 questions per LooGLE v2 document, the reader would otherwise often face an empty store. On LoCoMo and Mem-Gallery, earlier questions already write many experiences, and the drops are smaller (0.57 to 2.40).

\emph{Experience memory} (``MMMem'' \vs ``w/o Experience''). Removing experience memory lowers performance in all six settings. The largest drops occur on LooGLE v2 with Qwen3.5-9B (5.76 points) and MiniCPM-V~4.5 (2.47 points); on Mem-Gallery with Qwen3.5-9B, the drop is 2.09 points.
Since warm-up acts by writing experiences, this variant also loses warm-up. On LooGLE v2 with Qwen, disabling experience memory performs slightly better than disabling warm-up alone. Full MMMem outperforms both variants, supporting the benefit of the complete experience pathway.

\emph{Generated views} (``MMMem'' \vs ``w/o Generated views''). For visual retrieval, generated views combine clues from multiple records into a shared representation (Sec.~\ref{sec:views}). Both readers score higher with these views on LoCoMo and Mem-Gallery, supporting their role as an additional entry point to distributed source evidence. On the text-centric LooGLE v2 benchmark, models tend to skip scene generation and retrieve text directly. Overall, training and task warm-up contribute in every setting, whereas the size of the contributions of experience memory and generated views varies across benchmarks and readers.

\subsection{Case Study}
\label{sec:case_study}
\label{sec:trace}

Figure~\ref{fig:cases} presents three Mem-Gallery cases with MiniCPM-V~4.5, tracing how generated views, experience reuse, and task warm-up affect retrieval and answering.

For bird identification, descriptions of the target bird are spread across two records. A generated scene combines these descriptions into an image for visual matching, moving the relevant node from 10th to 2nd. Its source evidence then reaches the reader, and the answer changes from ``Blue Jay'' to ``Lazuli Bunting.'' Scene generation thus makes dispersed textual clues accessible through visual retrieval.

The counting question asks how many Dunlin photos Lin Wei shared. Both runs receive the same records and photos, including an earlier image. An experience from a previous scientific-name question links ``Dunlin'' to \emph{Calidris alpina}. With this relation available, the reader includes the earlier photo and changes the count from five to six. Experience helps connect species references within evidence already retrieved.

In the dog-breed case, the source record gives both the name ``Rocky'' and the breed ``Sussex Spaniel,'' and both runs cite it. Without warm-up, the recalled experience comes from a name question and supplies ``Rocky,'' which the reader incorrectly returns as the breed. With warm-up, the recalled experience concerns a therapy companion and supplies the breed cue ``Sussex,'' leading to ``Sussex Spaniel.'' This case illustrates how the task context in recalled experience can guide the reader toward the relevant information in a source. Warm-up thus prepares useful task experience in advance, helping later questions draw on knowledge relevant to their needs.

\section{Limitations}
\label{sec:limitations}

Experience reuse depends on whether earlier conclusions remain applicable as tasks and source information change. Future work could extend MMMem with mechanisms for revising and retiring experiences, allowing memory to track evolving knowledge over longer interactions.

Closer coordination between external consolidation and internal parameter adaptation is another direction. Training currently omits the dynamic summaries available to the experience writer at inference. Incorporating evolving memory states into training could better align experience formation with subsequent use.

Experience writing and retention introduce computation and storage costs alongside offline training. Selective writing, memory compression, and adaptive retrieval budgets offer promising directions for balancing answer quality, latency, and memory use in sustained interactions.

\section{Conclusion}
\label{sec:conclusion}

We introduced \textbf{MMMem}, a task-aware multimodal memory framework that combines external experience consolidation with internal parameter adaptation. External consolidation turns source knowledge and task conclusions into reusable experiences, enabling agents to retrieve, reuse, and retain task-relevant knowledge. Internal adaptation trains a shared model to answer from multimodal evidence and distill task experience, aligning its parametric knowledge with task demands. Source-generated tasks connect these processes through memory warm-up and self-training. Experiments across three memory benchmarks, supported by ablations and case studies, show the benefits of jointly improving memory organization and the model's ability to use accumulated knowledge.

\subsection*{Reproducibility statement}
The evaluation settings and self-training procedure are described in Sections~\ref{sec:setup} and~\ref{sec:selfgen}. Reference answers and annotated clues are excluded from retrieval, and failed or malformed outputs remain in the evaluation denominator.

\section*{LLM Usage Statement}
OpenAI Codex assisted with literature organization, drafting and language editing, figure planning, and analysis of experimental results. The authors are responsible for the content of the final manuscript.

{
  \small
  \bibliographystyle{ieeenat_fullname}
  \bibliography{references}

@inproceedings{asai2024selfrag,
  title = {{Self-RAG}: Learning to Retrieve, Generate, and Critique through Self-Reflection},
  author = {Asai, Akari and Wu, Zeqiu and Wang, Yizhong and Sil, Avirup and Hajishirzi, Hannaneh},
  booktitle = {International Conference on Learning Representations},
  year = {2024}
}

@inproceedings{faysse2025colpali,
  title = {{ColPali}: Efficient Document Retrieval with Vision Language Models},
  author = {Faysse, Manuel and Sibille, Hugues and Wu, Tony and Omrani, Bilel and Viaud, Gautier and Hudelot, C{\'e}line and Colombo, Pierre},
  booktitle = {International Conference on Learning Representations},
  year = {2025}
}

@inproceedings{hu2022lora,
  title = {{LoRA}: Low-Rank Adaptation of Large Language Models},
  author = {Hu, Edward J. and Shen, Yelong and Wallis, Phillip and Allen-Zhu, Zeyuan and Li, Yuanzhi and Wang, Shean and Wang, Lu and Chen, Weizhu},
  booktitle = {International Conference on Learning Representations},
  year = {2022}
}

@inproceedings{lewis2020rag,
  title = {Retrieval-Augmented Generation for Knowledge-Intensive {NLP} Tasks},
  author = {Lewis, Patrick and Perez, Ethan and Piktus, Aleksandra and Petroni, Fabio and Karpukhin, Vladimir and Goyal, Naman and K{\"u}ttler, Heinrich and Lewis, Mike and Yih, Wen-tau and Rockt{\"a}schel, Tim and Riedel, Sebastian and Kiela, Douwe},
  booktitle = {Advances in Neural Information Processing Systems},
  volume = {33},
  year = {2020}
}

@article{li2026ocrmemory,
  title = {{OCR-Memory}: Optical Context Retrieval for Long-Horizon Agent Memory},
  author = {Li, Jinze and Zhang, Yang and Yang, Xin and Qu, Jiayi and Xu, Jinfeng and Yang, Shuo and Ding, Junhua and Ngai, Edith Cheuk-Han},
  journal = {arXiv preprint arXiv:2604.26622},
  year = {2026}
}

@article{shi2026memocr,
  title = {{MemOCR}: Layout-Aware Visual Memory for Efficient Long-Horizon Reasoning},
  author = {Shi, Yaorui and Liu, Shugui and Yang, Yu and Mao, Wenyu and Chen, Yuxin and Gu, Qi and Su, Hui and Cai, Xunliang and Wang, Xiang and Zhang, An},
  journal = {arXiv preprint arXiv:2601.21468},
  year = {2026}
}

@inproceedings{xu2024recomp,
  title = {{RECOMP}: Improving Retrieval-Augmented {LM}s with Compression and Selective Augmentation},
  author = {Xu, Fangyuan and Shi, Weijia and Choi, Eunsol},
  booktitle = {International Conference on Learning Representations},
  year = {2024}
}

@inproceedings{yu2025visrag,
  title = {{VisRAG}: Vision-Based Retrieval-Augmented Generation on Multi-Modality Documents},
  author = {Yu, Shi and Tang, Chaoyue and Xu, Bokai and Cui, Junbo and Ran, Junhao and Yan, Yukun and Liu, Zhenghao and Wang, Shuo and Han, Xu and Liu, Zhiyuan and Sun, Maosong},
  booktitle = {International Conference on Learning Representations},
  year = {2025}
}

@inproceedings{maharana2024locomo,
  title = {Evaluating Very Long-Term Conversational Memory of {LLM} Agents},
  author = {Maharana, Adyasha and Lee, Dong-Ho and Tulyakov, Sergey and Bansal, Mohit and Barbieri, Francesco and Fang, Yuwei},
  booktitle = {Proceedings of the 62nd Annual Meeting of the Association for Computational Linguistics},
  pages = {13851--13870},
  year = {2024}
}

@inproceedings{he2025looglev2,
  title = {{LooGLE v2}: Are {LLM}s Ready for Real World Long Dependency Challenges?},
  author = {He, Ziyuan and Wang, Yuxuan and Li, Jiaqi and Liang, Kexin and Zhang, Muhan},
  booktitle = {Advances in Neural Information Processing Systems, Datasets and Benchmarks Track},
  year = {2025}
}

@article{chhikara2025mem0,
  title = {{Mem0}: Building Production-Ready {AI} Agents with Scalable Long-Term Memory},
  author = {Chhikara, Prateek and Khant, Dev and Aryan, Saket and Singh, Taranjeet and Yadav, Deshraj},
  journal = {arXiv preprint arXiv:2504.19413},
  year = {2025}
}

@inproceedings{xu2025amem,
  title = {{A-Mem}: Agentic Memory for {LLM} Agents},
  author = {Xu, Wujiang and Liang, Zujie and Mei, Kai and Gao, Hang and Tan, Juntao and Zhang, Yongfeng},
  booktitle = {Advances in Neural Information Processing Systems},
  year = {2025}
}

@article{bei2026memgallery,
  title = {{Mem-Gallery}: Benchmarking Multimodal Long-Term Conversational Memory for {MLLM} Agents},
  author = {Bei, Yuanchen and Wei, Tianxin and Ning, Xuying and Zhao, Yanjun and Liu, Zhining and Lin, Xiao and Zhu, Yada and Hamann, Hendrik and He, Jingrui and Tong, Hanghang},
  journal = {arXiv preprint arXiv:2601.03515},
  year = {2026}
}

@inproceedings{shinn2023reflexion,
  title = {Reflexion: Language Agents with Verbal Reinforcement Learning},
  author = {Shinn, Noah and Cassano, Federico and Berman, Edward and Gopinath, Ashwin and Narasimhan, Karthik and Yao, Shunyu},
  booktitle = {Advances in Neural Information Processing Systems},
  year = {2023}
}

@inproceedings{zhao2024expel,
  title = {{ExpeL}: {LLM} Agents Are Experiential Learners},
  author = {Zhao, Andrew and Huang, Daniel and Xu, Quentin and Lin, Matthieu and Liu, Yong-Jin and Huang, Gao},
  booktitle = {Proceedings of the AAAI Conference on Artificial Intelligence},
  year = {2024}
}

@article{wang2024awm,
  title = {Agent Workflow Memory},
  author = {Wang, Zora Zhiruo and Mao, Jiayuan and Fried, Daniel and Neubig, Graham},
  journal = {arXiv preprint arXiv:2409.07429},
  year = {2024}
}

@inproceedings{ouyang2026reasoningbank,
  title = {{ReasoningBank}: Scaling Agent Self-Evolving with Reasoning Memory},
  author = {Ouyang, Siru and Yan, Jun and Hsu, I-Hung and Chen, Yanfei and Jiang, Ke and Wang, Zifeng and Han, Rujun and Le, Long T. and Daruki, Samira and Tang, Xiangru and Tirumalashetty, Vishy and Lee, George and Rofouei, Mahsan and Lin, Hangfei and Han, Jiawei and Lee, Chen-Yu and Pfister, Tomas},
  booktitle = {International Conference on Learning Representations},
  year = {2026}
}

@inproceedings{zelikman2022star,
  title = {{STaR}: Bootstrapping Reasoning With Reasoning},
  author = {Zelikman, Eric and Wu, Yuhuai and Mu, Jesse and Goodman, Noah},
  booktitle = {Advances in Neural Information Processing Systems},
  volume = {35},
  pages = {15476--15488},
  year = {2022},
  doi = {10.52202/068431-1126},
  url = {https://proceedings.neurips.cc/paper_files/paper/2022/hash/639a9a172c044fbb64175b5fad42e9a5-Abstract-Conference.html},
  eprint = {2203.14465},
  archivePrefix = {arXiv}
}

@article{gulcehre2023rest,
  title = {Reinforced Self-Training ({ReST}) for Language Modeling},
  author = {Gulcehre, Caglar and Paine, Tom Le and Srinivasan, Srivatsan and Konyushkova, Ksenia and Weerts, Lotte and Sharma, Abhishek and Siddhant, Aditya and Ahern, Alex and Wang, Miaosen and Gu, Chenjie and Macherey, Wolfgang and Doucet, Arnaud and Firat, Orhan and de Freitas, Nando},
  journal = {arXiv preprint arXiv:2308.08998},
  year = {2023},
  eprint = {2308.08998},
  archivePrefix = {arXiv},
  url = {https://arxiv.org/abs/2308.08998}
}

@inproceedings{hu2024vpd,
  title     = {Visual Program Distillation: Distilling Tools and Programmatic Reasoning into Vision-Language Models},
  author    = {Hu, Yushi and Stretcu, Otilia and Lu, Chun-Ta and Viswanathan, Krishnamurthy and Hata, Kenji and Luo, Enming and Krishna, Ranjay and Fuxman, Ariel},
  booktitle = {Proceedings of the IEEE/CVF Conference on Computer Vision and Pattern Recognition (CVPR)},
  year      = {2024},
  eprint    = {2312.03052},
  archivePrefix = {arXiv}
}

@article{jiang2026delivermem,
  title = {Retrieved but Not Delivered: Multimodal Memory Delivery for Long-Term Agents},
  author = {Jiang, Yuhang and Liao, Qingwei and Yin, Kaize and Liu, Xingling and Cuomo, Luca and Bacci, Silvio},
  journal = {arXiv preprint arXiv:2609.32590},
  year = {2026}
}

@article{gao2026memo,
  title = {{MEMO}: Multimodal Evidence Memory Organization for Long-Horizon {LLM} Agents},
  author = {Gao, Xian and Wang, Jinpeng and Ruan, Jiacheng and Cao, Guangyu and Liu, Ting and Fu, Yuzhuo},
  journal = {arXiv preprint arXiv:2609.07471},
  year = {2026}
}

@inproceedings{petroni2019lama,
  title = {Language Models as Knowledge Bases?},
  author = {Petroni, Fabio and Rockt{\"a}schel, Tim and Riedel, Sebastian and Lewis, Patrick and Bakhtin, Anton and Wu, Yuxiang and Miller, Alexander},
  booktitle = {Proceedings of the 2019 Conference on Empirical Methods in Natural Language Processing and the 9th International Joint Conference on Natural Language Processing (EMNLP-IJCNLP)},
  year = {2019},
  pages = {2463--2473},
  doi = {10.18653/v1/D19-1250},
  url = {https://aclanthology.org/D19-1250/}
}

@article{zhang2024raft,
  title = {{RAFT}: Adapting Language Model to Domain Specific {RAG}},
  author = {Zhang, Tianjun and Patil, Shishir G. and Jain, Naman and Shen, Sheng and Zaharia, Matei and Stoica, Ion and Gonzalez, Joseph E.},
  journal = {arXiv preprint arXiv:2403.10131},
  year = {2024},
  eprint = {2403.10131},
  archivePrefix = {arXiv},
  url = {https://arxiv.org/abs/2403.10131}
}

@inproceedings{zeng2025kbalign,
  title = {{KBAlign}: Efficient Self Adaptation on Specific Textual Knowledge Bases},
  author = {Zeng, Zheni and Chen, Yuxuan and Yu, Shi and Wang, Ruobing and Yan, Yukun and Liu, Zhenghao and Wang, Shuo and Han, Xu and Liu, Zhiyuan and Sun, Maosong},
  booktitle = {Findings of the Association for Computational Linguistics: EMNLP 2025},
  year = {2025},
  pages = {13519--13532},
  doi = {10.18653/v1/2025.findings-emnlp.728},
  url = {https://aclanthology.org/2025.findings-emnlp.728/}
}

@article{behrouz2026sleep,
  title = {Language Models Need Sleep: Learning to Self-Modify and Consolidate Memories},
  author = {Behrouz, Ali and Hashemi, Farnoosh and Javanmard, Adel and Mirrokni, Vahab},
  journal = {arXiv preprint arXiv:2606.03979},
  year = {2026},
  url = {https://arxiv.org/abs/2606.03979}
}

@article{ren2026tmem,
  title = {Scaling Self-Evolving Agents via Parametric Memory},
  author = {Ren, Tao and Luo, Weiyao and Yang, Hui and Zhu, Rongzhi and Huang, Xiang and Wu, Yuchuan and Chou, Bingxue and Ye, Jieping and Liang, Jiafeng and Li, Yongbin and Peng, Yijie},
  journal = {arXiv preprint arXiv:2606.04536},
  year = {2026},
  url = {https://arxiv.org/abs/2606.04536}
}

@article{ye2026oel,
  title = {Online Experiential Learning for Language Models},
  author = {Ye, Tianzhu and Dong, Li and Dong, Qingxiu and Wu, Xun and Huang, Shaohan and Wei, Furu},
  journal = {arXiv preprint arXiv:2603.16856},
  year = {2026},
  url = {https://arxiv.org/abs/2603.16856}
}

@article{xia2026unimem,
  title = {{UniMem}: Complementary Episodic-to-Parametric Memory for Boundary-Agnostic Task Streams},
  author = {Xia, Siyu and Zhang, Chenheng and Wu, Yanting and Li, Haoxuan and Chai, Jiajun and Wang, Xiaohan and Yin, Guojun and Lin, Wei and Lin, Zhouchen and Zhang, Haifeng and Wang, Jun},
  journal = {arXiv preprint arXiv:2607.26017},
  year = {2026},
  url = {https://arxiv.org/abs/2607.26017}
}

@article{wang2024lvbench,
  title = {{LVBench}: An Extreme Long Video Understanding Benchmark},
  author = {Wang, Weihan and He, Zehai and Hong, Wenyi and Cheng, Yean and Zhang, Xiaohan and Qi, Ji and Gu, Xiaotao and Huang, Shiyu and Xu, Bin and Dong, Yuxiao and Ding, Ming and Tang, Jie},
  journal = {arXiv preprint arXiv:2406.08035},
  year = {2024},
  url = {https://arxiv.org/abs/2406.08035}
}

@article{li2025ovobench,
  title = {{OVO-Bench}: How Far is Your {Video-LLMs} from Real-World Online Video Understanding?},
  author = {Li, Yifei and Niu, Junbo and Miao, Ziyang and Ge, Chunjiang and Zhou, Yuanhang and He, Qihao and Dong, Xiaoyi and Duan, Haodong and Ding, Shuangrui and Qian, Rui and Zhang, Pan and Zang, Yuhang and Cao, Yuhang and He, Conghui and Wang, Jiaqi},
  journal = {arXiv preprint arXiv:2501.05510},
  year = {2025},
  url = {https://arxiv.org/abs/2501.05510}
}
}

\clearpage
\appendix
\twocolumn[{%
  \centering
  \pdfbookmark[0]{Appendix}{mmmem-appendix}
  {\Large\bfseries Appendix\par}
  \vspace{0.6em}
}]
\section{Evaluation Protocol}
\label{sec:exp_appendix}
\label{sec:shared-experimental-setting}

\subsection{Evaluation Units and Scoring}
Table~\ref{tab:training_pools} gives the held-out evaluation subsets, source-view budgets, and eligible training pools. A source unit is a complete dialogue in LoCoMo, a scenario in Mem-Gallery, and a document in LooGLE v2. A question can draw on the history of its associated unit, including earlier sessions of the same dialogue or scenario. Reference answers, annotated clue spans, and QA tags are excluded from the retrieval index.

\smallskip
\noindent
\begin{minipage}{\columnwidth}
\ifdefined\nolinenumbers\nolinenumbers\fi
\centering
\small
\setlength{\tabcolsep}{3pt}
\begin{tabular*}{\linewidth}{@{\extracolsep{\fill}}lrrrr@{}}
\toprule
Benchmark & Eval. units & Questions & Tokens & Train units \\
\midrule
LoCoMo~\cite{maharana2024locomo} & 6 & 1,178 & 3,000 & 2 \\
Mem-Gallery~\cite{bei2026memgallery} & 12 & 1,116 & 4,000 & 4 \\
LooGLE v2~\cite{he2025looglev2} & 766 & 1,164 & 4,000 & 358 \\
LVBench~\cite{wang2024lvbench} & --- & --- & --- & 10 \\
OVO-Bench~\cite{li2025ovobench} & --- & --- & --- & 138 \\
\midrule
Total training pool & & & & 512 \\
\bottomrule
\end{tabular*}
\captionsetup{hypcap=false}
\captionof{table}{\textbf{Evaluation subsets and training source pools.} Tokens are the per-question source-view budget for MMMem. Train units count eligible sources before task sampling. Video sources from LVBench and OVO-Bench contribute to training.}
\label{tab:training_pools}
\end{minipage}
\par\smallskip

We report token F1 on LoCoMo and exact match, token F1, and BLEU-1 on Mem-Gallery. Mem-Gallery uses \textit{Not mentioned.} as the reference response for unanswerable questions. LooGLE v2 uses its official aggregate score, including partial credit for version-control questions. The denominator includes failed and malformed responses.

\subsection{Baseline Implementations}
\label{sec:appendix-baselines}
Any baseline memory shared across scored questions is built from the unit's source history. Memory created for an individual question is used only for that question.

\paragraph{Context and dense retrieval.}
Direct Answer uses the question and any accompanying query image. Full Context supplies up to 32K history tokens. Recent Context supplies the most recent 3K history tokens on LoCoMo and 4K on Mem-Gallery and LooGLE v2. Dense RAG retrieves historical passages from a segment index by semantic similarity.

\paragraph{Visual memory.}
OCR-Memory selects historical material from rendered memory pages for the current question. MemOCR starts with an empty draft, writes a memory from the eight highest-ranked history passages, and renders the draft as an image for the reader.

\paragraph{Agent memory.}
Mem0 builds a fact store from the historical sessions before evaluation. On LoCoMo, it maintains a separate store for each of the two speakers. At evaluation it searches the stored facts for each question. A-Mem adds the historical turns as linked notes during construction and uses keywords from the current question to retrieve notes for the reader.

\subsection{Settings for the Ablation Study}
\paragraph{Parameter adaptation.}
\emph{Only Training} is Dense RAG with the same adapted answering checkpoint as full MMMem. It uses segment retrieval without the MMMem warm-up and experience workflow. The comparison with Dense RAG measures the effect of adaptation within the same retrieval setup. The comparison with full MMMem holds the adapted answering checkpoint fixed. \emph{w/o Training} uses the original backbone in the MMMem workflow. Full MMMem and this variant use the same warm-up questions within each backbone setting.

\paragraph{Memory components.}
\emph{w/o Generated views} removes the generated scenes from the source memory while retaining the other view types. The results for this variant on LooGLE v2 are unreported. \emph{w/o Task warm-up} starts scored evaluation with an empty experience store and accumulates experience as questions are answered. \emph{w/o Experience} disables the experience store and its write, recall, and retention operations, including the warm-up that populates it.

\section{MMMem Implementation}
\label{sec:appendix-implementation}

\subsection{Multimodal Memory Construction}
\label{sec:appendix-construction}
\paragraph{Source nodes.}
The memory assigns a persistent ID to each source record. Within a session, consecutive turns are packed into segments of up to six turns and 400 tokens. A final segment of fewer than three turns is merged into the previous segment when the result fits those limits. A turn longer than the token limit forms a segment of its own. Each session node points to its segments. Entity nodes group records with a shared capitalized name appearing at least twice, with speaker names and sentence-initial tokens excluded from name extraction. One timeline node per unit groups records containing temporal expressions.

\paragraph{Scene views.}
Entity and timeline nodes are eligible for scene generation at construction. A node used by three tasks can also become eligible, including a segment node. Scene generation is capped at 48 views per unit. The specification prompt in Section~\ref{sec:prompt-scenes} receives all source records in the selected node and selects facts and their record IDs for one scene. We keep facts whose IDs belong to the node. Qwen-Image renders the scene description, and a vision-language check compares the number and genders of people and the setting with the specification. A failed specification or image check is retried once. The accepted view retains the source IDs of the depicted facts.

\subsection{Self-Improving Training}
\label{sec:appendix-task-generation}
\paragraph{Task generation.}
The generator writes questions, target answers, and supporting record IDs from the supplied source records. The templates in Section~\ref{sec:prompt-synthesis} cover free-answer and multiple-choice questions, including related sequences that revisit a fact or ask for new information from the same source. For basic QA generation, the prefix contains six official question texts sampled from the same domain's training pool as style examples. The pool can include questions from the training unit being processed. Generation relies on the prompt to elicit new wording. Image-based QA is generated directly from the source records.

\paragraph{Warm-up as external experience consolidation.}
Before the first scored question, MMMem initializes an experience store for the evaluation unit and generates unscored tasks from its source material. The style examples come from the separate training pool. For each warm-up task, the reader takes the generated question, any options, and retrieved evidence. A second call consolidates the reader's prediction and source evidence into an experience available to scored questions. Task generation and these per-task answering and writing calls are part of the test-time workload.

\paragraph{Warm-up question overlap.}
We compare warm-up and scored question texts within each unit after Unicode NFKC normalization, case folding, and whitespace collapse. On LoCoMo, 2 of 887 Qwen warm-up questions and 10 of 843 MiniCPM warm-up questions have exact matches. Both readers have zero exact matches on Mem-Gallery and LooGLE v2. Excluding the matched scored questions changes LoCoMo F1 from 65.84 to 65.82 for Qwen and from 56.40 to 56.27 for MiniCPM.

\paragraph{Self-training as parameter adaptation.}
Offline supervision is generated from training units disjoint from evaluation. The targets teach the model to answer with source IDs and to summarize a completed task as a conclusion linked to supporting records. Official training questions contribute style and abstention-format examples. For example, calibration assigns \textit{Not mentioned.} on a subset of Mem-Gallery training examples and \textit{Unable to answer} on a subset of OVO-Bench examples. The generator supplies the target answers, while official reference answers are reserved for evaluation. MiniCPM's training includes 54 image-based QA examples generated by Qwen3.5-9B.

\paragraph{Optimization.}
Each backbone is adapted separately. We apply LoRA to the language model's query, key, value, and output attention projections with rank 16, \(\alpha=32\), and dropout 0.05. We train for one epoch using AdamW at a learning rate of \(10^{-4}\), with a microbatch of one and eight gradient accumulation steps. Loss is computed on target tokens. The Qwen and MiniCPM checkpoints used for the main experiments are \texttt{JFIXQ\_train\_ids} and \texttt{JFIXM\_train\_teacher}, respectively.

\subsection{Task-Aware Online Editing}
\label{sec:appendix-online}
\paragraph{Source retrieval and reader inputs.}
LoCoMo and Mem-Gallery use text and visual retrieval, while retrieval on LooGLE v2 is predominantly text-based. In Mem-Gallery, historical images are indexed by both their captions and their visual representations. A query image contributes to rank fusion and is supplied to the reader. We rerank candidates with the pretrained Qwen3-Reranker-0.6B cross-encoder in a zero-shot setting and select source evidence within the token budget in Table~\ref{tab:training_pools}. A match to a generated scene recovers the corresponding node's source text and original images. A separate slot supplies the top-ranked staged experience by question-text relevance.

\paragraph{Scored questions and experience writing.}
After warm-up, MMMem answers scored questions in dataset order using fixed reader and writer parameters. The answering call receives the current task, retrieved evidence, and any recalled experience. The experience-writing call uses the model's own answer and the source records it cited. Each new experience stores the task context, conclusion, and supporting record IDs and becomes available to subsequent questions in the same unit. Benchmark reference answers and scoring feedback are kept separate from the reader and writer. Section~\ref{sec:prompt-reader} gives the reader prompts, and Section~\ref{sec:prompt-writer} gives the summary prompt.

\paragraph{Reuse within an evaluation unit.}
An experience receives adoption credit when it reaches the reader and the answer cites a linked record that supports it. Its first adoption can trigger an associated image. Rephrasings of a question contribute up to two adoptions. The system checks the 32-entry staging area every eight tasks or when it fills. It promotes experiences with at least two adoptions into the source index and expires staged entries that go 32 tasks without adoption. At the next dialogue, scenario, or document, the system initializes a fresh experience store.

\lstdefinestyle{mmmem-prompt}{
  basicstyle=\ttfamily\footnotesize,
  columns=fullflexible,
  keepspaces=true,
  showstringspaces=false,
  upquote=true,
  breaklines=true,
  breakatwhitespace=true,
  breakautoindent=false,
  breakindent=0pt,
  frame=single,
  framesep=5pt,
  rulecolor=\color{black!35},
  backgroundcolor=\color{black!3},
  xleftmargin=6pt,
  xrightmargin=6pt,
  aboveskip=5pt,
  belowskip=5pt
}
\newenvironment{mmmpromptblock}{%
  \par\noindent\begin{minipage}{\linewidth}\ifdefined\nolinenumbers\nolinenumbers\fi
}{\end{minipage}\par\smallskip}
\lstnewenvironment{mmmprompt}[1]{%
  \lstset{style=mmmem-prompt,title={\small\bfseries #1}}%
}{}

\begin{mmmpromptblock}
\section{Prompts}
\label{sec:appendix-prompts}

\subsection{Scene-View Generation}
\label{sec:prompt-scenes}

The returned \texttt{scene} string is passed directly to Qwen-Image. The image check extracts visible attributes for comparison with the specification.

\begin{mmmprompt}{Scene specification}
You design one illustration that helps someone recall a memory. Read the records and return JSON with keys: facts (list of {text, source_id}; only facts stated in the records), entities (list of names), scene (one sentence describing a single picture of the most memorable content: who, doing what, where; no dates, no numbers, no invented details), people (integer number of people in the picture), genders (list with one entry per person, each 'man', 'woman' or 'unknown', inferred only from the records, e.g. pronouns or names used), setting (one or two words, e.g. 'dance studio', 'beach'). The scene must describe people by gender (e.g. 'a man'), never by name alone. Return JSON only.

Records:
{records}
\end{mmmprompt}
\end{mmmpromptblock}

\begin{mmmpromptblock}
\begin{mmmprompt}{Image check}
Look at the image and report only what is visible, as JSON: {"people": <integer count of people>, "genders": ["man" | "woman" | "unknown", one per person], "setting": "<one or two words for the place>", "objects": ["<up to 6 main objects>"]}. Return JSON only.
\end{mmmprompt}
\end{mmmpromptblock}

\subsection{Synthetic QA Generation}
\label{sec:prompt-synthesis}

The templates use \texttt{\{note\}} to describe attached photos or video keyframes when applicable.

\begin{mmmpromptblock}
\begin{mmmprompt}{Few-shot prefix}
Here are real questions users asked about other material of this kind. Match their style, difficulty and mix of question types, but write new questions about the records below; never copy these:
{examples}
\end{mmmprompt}
\end{mmmpromptblock}

\begin{mmmpromptblock}
\begin{mmmprompt}{Independent free-answer questions}
{note}Below are conversation records, each starting with [id] speaker (date). Write {n} question-answer pairs a person could later ask about this conversation. Mix these types: single_hop (one fact), temporal (when something happened; answer with an absolute date or month), multi_hop (needs two records), no_information (plausible question the records do NOT answer; answer exactly 'No information available' and give a wrong but plausible alternative). Answers must be short. Return JSON: [{"type": str, "question": str, "answer": str, "evidence": [ids], "distractor": str (only for no_information)}]. Return JSON only.

Records:
{records}
\end{mmmprompt}
\end{mmmpromptblock}

\begin{mmmpromptblock}
\begin{mmmprompt}{Independent multiple-choice questions}
{note}Below are records, each starting with [id]. Write {n} multiple-choice questions a person could later ask about this material. Each question has exactly four options; exactly one is correct and the other three are plausible but contradicted or unsupported by the records. Mix: detail (one fact), order/time, count or comparison, reason. Return JSON: [{"type": str, "question": str, "options": [4 strings], "answer_index": 0-3, "evidence": [ids]}]. Return JSON only.

Records:
{records}
\end{mmmprompt}
\end{mmmpromptblock}

\begin{mmmpromptblock}
\begin{mmmprompt}{Free-answer task sequence}
{note}Below are conversation records, each starting with [id] speaker (date). Write one related task sequence of four questions a user might ask at different times about the SAME records:
1. fact: one fact; 2. paraphrase: the same fact asked differently; 3. other_need: a different fact or a comparison from these records; 4. near_miss: like question 1 but with a different person or date so the records do NOT answer it (answer exactly 'No information available' and give a plausible wrong alternative as distractor).
Return JSON: [{"step": "fact"|"paraphrase"|"other_need"|"near_miss", "question": str, "answer": str, "evidence": [ids], "distractor": str (near_miss only)}]. Return JSON only.

Records:
{records}
\end{mmmprompt}
\end{mmmpromptblock}

\begin{mmmpromptblock}
\begin{mmmprompt}{Multiple-choice task sequence}
{note}Below are records, each starting with [id]. Write one related sequence of three multiple-choice questions about the SAME records: 1. fact; 2. paraphrase of the same fact with reordered options; 3. other_need: a different fact, order or count from these records. Four options each, one correct.
Return JSON: [{"step": "fact"|"paraphrase"|"other_need", "question": str, "options": [4 strings], "answer_index": 0-3, "evidence": [ids]}]. Return JSON only.

Records:
{records}
\end{mmmprompt}
\end{mmmpromptblock}

\subsection{Reader}
\label{sec:prompt-reader}

Braced fields are filled at runtime. The option instruction, task constraints, and query-image note are included when applicable. The query image is the first image attachment when present. Retrieved text and any recalled experience appear under \texttt{Memory}.

\begin{mmmpromptblock}
\begin{mmmprompt}{Grounded answering}
Answer the question using only the memory below. Each record starts with [id] speaker (date). Memory notes marked 'unverified' are your own earlier answers; trust the records over them.
Reply in exactly this format:
Answer: <short answer>
Evidence: <comma-separated record ids you used>
If the memory does not contain enough information, reply instead:
Answer: insufficient
Need: <what to look up next>
{option_instruction}
{task_constraints}

Question: {public_query}
{query_image_note}

Memory:
{memory}
\end{mmmprompt}
\end{mmmpromptblock}

\begin{mmmpromptblock}
\begin{mmmprompt}{Option instruction}
The question lists options (a), (b), ...: answer with the text of exactly one option.
\end{mmmprompt}
\end{mmmpromptblock}

\begin{mmmpromptblock}
\begin{mmmprompt}{Query-image note}
(The first image is the picture shown with the question.)
\end{mmmprompt}
\end{mmmpromptblock}

For scored LooGLE v2 questions, we fill the benchmark-provided instruction with the retrieved context, options, and question, then append the suffix below. For LooGLE v2 warm-up, we use the context template below with the same suffix.

\begin{mmmpromptblock}
\begin{mmmprompt}{LooGLE v2 warm-up template}
Answer the question using only the context.

Context:
{context}

{options}Question: {question}
\end{mmmprompt}
\end{mmmpromptblock}

\begin{mmmpromptblock}
\begin{mmmprompt}{LooGLE v2 answer suffix}
Do not explain. Your entire response must be one line beginning exactly `The correct answer is ` followed only by the final answer requested above.
\end{mmmprompt}
\end{mmmpromptblock}

\subsection{Experience Writing}
\label{sec:prompt-writer}

The writer produces the source-cited summary. The runner stores it with the task description and the linked source records.

\begin{mmmpromptblock}
\begin{mmmprompt}{Experience summary}
You just answered a question from memory. In at most two sentences, summarise what the memory says that answers it (or that it says nothing about it), citing record ids in brackets. Do not add facts that are not in the records.

Question: {question}
Your answer: {answer}

Records you read:
{records}

Summary:
\end{mmmprompt}
\end{mmmpromptblock}

\twocolumn[{
\begin{minipage}{\textwidth}
\ifdefined\nolinenumbers\nolinenumbers\fi
\section{Comparison with Related Memory Methods}
\label{sec:appendix-memory-comparison}

Table~\ref{tab:compare_full} compares the original methods by the memory they keep, how they retrieve and update it, and the components they train. Section~\ref{sec:appendix-baselines} describes how we run the baselines in our experiments.

\medskip
\begingroup
  \centering
  \small
  \setlength{\tabcolsep}{3pt}
  \renewcommand{\arraystretch}{1.12}
  \begin{tabular*}{\textwidth}{@{\extracolsep{\fill}}>{\raggedright\arraybackslash}p{2.7cm}>{\raggedright\arraybackslash}p{3.25cm}>{\raggedright\arraybackslash}p{2.5cm}>{\raggedright\arraybackslash}p{4.0cm}>{\raggedright\arraybackslash}p{2.45cm}@{}}
    \toprule
    Method & Memory content & Retrieval / reading & Memory updates & Trained component \\
    \midrule
    Mem0~\cite{chhikara2025mem0}
      & Extracted facts & Text search & Add, revise, or delete facts & -- \\
    A-Mem~\cite{xu2025amem}
      & Linked text notes & Text search & Link and revise notes & -- \\
    OCR-Memory~\cite{li2026ocrmemory}
      & Rendered pages linked to source text & Visual search & Refresh recalled content at higher resolution & Retriever \\
    MemOCR~\cite{shi2026memocr}
      & Rewritten memory pages & Page reading & Rewrite pages within a budget & Reader/writer \\
    DeliverMem~\cite{jiang2026delivermem}
      & Text and original images & Text and visual search & Keep records with identifiers and time information & -- \\
    MEMO~\cite{gao2026memo}
      & Source-linked evidence & Text search & Select and format evidence for each query & Extractor, manager \\
    Reflexion~\cite{shinn2023reflexion}
      & Task reflections & Recent reflections & Keep reflections in a bounded window & -- \\
    ExpeL~\cite{zhao2024expel}
      & Trajectories and task insights & Text search & Edit and vote on insights & -- \\
    AWM~\cite{wang2024awm}
      & Reusable workflows & Workflows in the prompt & Add induced workflows & -- \\
    ReasoningBank~\cite{ouyang2026reasoningbank}
      & Trajectories and task insights & Text search & Add task-derived memory entries & -- \\
    \midrule
    \textbf{MMMem}
      & Text, images, generated scenes, and task experiences
      & Text and visual search
      & Write experiences; retain them through source-supported reuse
      & Reader/writer$^{\dagger}$ \\
    \bottomrule
  \end{tabular*}
  \captionsetup{hypcap=false}
  \captionof{table}{\textbf{Comparison with related memory methods.} Trained components refer to additional parameter training reported by the original methods; -- indicates none reported. $^{\dagger}$MiniCPM training includes 54 image-QA examples generated by Qwen3.5-9B.}
  \label{tab:compare_full}
\endgroup

\end{minipage}
\vspace{1em}
}]

\end{document}